\documentclass[11pt]{article}
\usepackage[a4paper,bottom=2cm,top=2cm,left=2.54cm,right=2.54cm]{geometry}

\usepackage[utf8]{inputenc} 
\usepackage[T1]{fontenc}    
\usepackage{hyperref}       
\usepackage{url}            
\usepackage{booktabs}       
\usepackage{amsfonts}       
\usepackage{amsmath}        
\usepackage{xcolor}         
\usepackage{graphicx}       
\usepackage{subfigure}      
\usepackage{multirow}       
\usepackage[noabbrev]{cleveref}       
\usepackage{nicefrac}       
\usepackage{microtype}      
\DeclareMathOperator{\argmin}{argmin}
\usepackage{caption}
\usepackage{environ}
\newcommand{\acksection}{\section*{Acknowledgments}}
\NewEnviron{ack}{%
  \acksection
  \BODY
}

\title{\bf Deeply Learned Spectral Total Variation Decomposition}

\author{%
  Tamara G. Grossmann\thanks{Department of Applied Mathematics and Theoretical Physics, University of Cambridge, Cambridge, UK \hspace*{1.8em} \texttt{\{tg410,yk362,cbs31\}@cam.ac.uk}}
  \and
  Yury Korolev\footnotemark[1]
  \and
  Guy Gilboa\thanks{Department of Electrical Engineering, Technion - Israel Institute of Technology,  Haifa, Israel \hspace*{1.8em}  \texttt{guy.gilboa@ee.technion.ac.il}} \and
  Carola-Bibiane Sch\"onlieb\footnotemark[1]
}
\date{}

\newcommand{\tg}[1]{\textcolor{black}{#1}}
\begin{document}

\maketitle

\begin{abstract}
  Non-linear spectral decompositions of images based on one-homogeneous functionals such as total variation have gained considerable attention in the last few years. Due to their ability to extract spectral components corresponding to objects of different size and contrast, such decompositions enable filtering, feature transfer, image fusion and other applications. However, obtaining this decomposition involves solving multiple non-smooth optimisation problems and is therefore computationally highly intensive. In this paper, we present a neural network approximation of a non-linear spectral decomposition. We report up to four orders of magnitude ($\times 10,000$) speedup in processing of mega-pixel size images, compared to classical GPU implementations. Our proposed network, TVSpecNET, is able to implicitly learn the underlying PDE and, despite being entirely data driven, inherits invariances of the model based transform. To the best of our knowledge, this is the first approach towards learning a non-linear spectral decomposition of images. Not only do we gain a staggering computational advantage, but this approach can also be seen as a step towards studying neural networks that can decompose an image into spectral components defined by a user rather than a handcrafted functional. 
\end{abstract}

\section{Introduction}
Transforming and processing information such as images in a frequency domain to facilitate analysis and manipulation is a classical and very successful approach. A prominent example of a linear spectral decomposition is the Fourier transform that uses the trigonometric basis to represent a signal or an image. However, this linear transform is not optimal for images that contain discontinuities (edges), which can only be represented using high frequencies. To overcome this, a non-linear spectral decomposition based on the edge-preserving total variation (TV) functional was proposed in \cite{Gilboa2013,Gilboa2014}. The TV transform enables a scale representation based on the size and contrast of the structures contained in an image. The spectral components are related to eigenfunctions induced by the TV functional such as  indicator functions of disks and other smooth convex shapes. Similarly to the linear case, the spectral components of an image defined by the non-linear TV transform can be filtered, extracted and attenuated at different scales. Manipulation and analysis of images using the spectral TV decomposition has found various successful applications ranging from image denoising \cite{Moeller2015} through texture extraction and separation \cite{Brox2004,Horesh2016} and image fusion \cite{Benning2017,Zhao2018,Hait2019,Liu2020} to non-linear segmentation in biomedical imaging \cite{Zeune2017}. \tg{Especially in image fusion applications, spectral TV decomposition is able to overcome challenges in relation to edge and detail preservation where other methods fail \cite{Zhao2018}.} The theory of non-linear spectral decomposition has been extended to arbitrary one-homogeneous functionals in \cite{Burger2015,Burger2016,bungert2019nonlinear} and $p$-homogeneous functionals ($p\in(1,2)$) in \cite{cohen2020pLapSpectra}. \tg{Recently, spectral TV decomposition has additionally been generalised from the Euclidean space to surfaces \cite{Fumero2020}.}

In order to obtain the spectral TV decomposition of an image, the solution to the TV flow needs to be computed at every scale. This involves solving multiple non-smooth optimisation problems and is therefore computationally costly. To overcome this issue, we consider training a neural network (NN) to reproduce the spectral TV decomposition at a considerably reduced computational cost. \tg{Since the TV transform can be seen as a non-linear analogue to the Fourier transform, we aim to obtain an analogue to the Fast Fourier Transform (FFT) in our approach. The availability of a fast computational method is arguably one of the reasons for the success of the Fourier transform in signal and image processing. Fast methods for non-linear spectral decomposition therefore have the potential to become an equally important tool for image and data analysis.}

Decomposing images into task dependent components via deep learning  has been used in multiple imaging tasks, such as denoising \cite{Zhang2017,Liu2018a,Zhang2018} (components are the noise-free image and noise), segmentation \cite{Ronneberger2015,Badrinarayanan2017,Gandelsman2019}, material decomposition \cite{Xu2018,Lu2019a,Li2020} (e.g. separation to  bones and soft-tissues in medical imaging) and intrinsic image decomposition \cite{Narihira2015,Janner2017,Lettry2018,Li2018} (shading  and reflectance). The task of obtaining a non-linear spectral decomposition is, however, different. While in the examples above, the components are defined semantically, i.e. they depend on the contents of the image, spectral TV decompositions are based on a PDE, hence to learn a spectral decomposition, the network has to implicitly learn a PDE. This allows one to apply the trained network to images significantly different from the training set. 

Training a NN that reproduces an analytical spectral decomposition based on a handcrafted functional is a first step towards an even more ambitious goal of learning user defined (data driven) decompositions. This would involve training a NN to reproduce the desired behaviour on user defined 'eigenfunctions', which can be then transferred to real images.

\begin{figure}
    \centering
    \includegraphics[width=\textwidth]{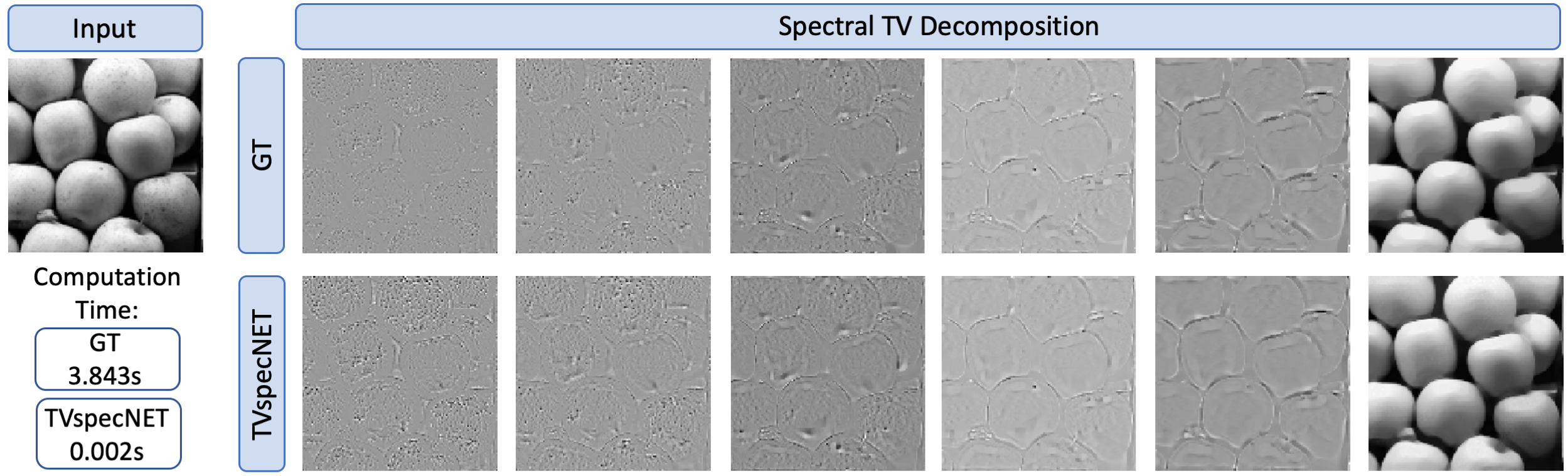}
    \caption{Visual comparison of our proposed TVspecNET decomposition and the ground truth (GT) \cite{Gilboa2014} on an example image from MS COCO \cite{Lin2014}. For this image, the resulting evaluation measures are: SSIM: 0.9849, PSNR: 32.046, sLMSE: 0.6062.}
    \label{fig.visual_comp_leo}
\end{figure}

\paragraph{Contributions}
In this paper, we propose a neural network that we call TVspecNET that can reproduce the spectral TV decomposition of images while significantly (by more than three orders of magnitude) reducing the computation time to obtain the decomposition once the network is trained. 
Our main contributions are as follows 
\begin{itemize}
    \item \tg{We approximate a highly non-trivial function by means of deep learning to obtain the non-linear spectral decomposition where model-driven approaches have been cumbersome and computationally complex;}
    \item We achieve a \tg{substantial} computational speed up compared to the classical,  model driven approach that is based on solving a gradient flow;
    \item We demonstrate that our network is indeed capable of learning intrinsic properties of the non-linear spectral decomposition such as one-homogeneity and rotational and translational invariance. Moreover, we demonstrate that the network not only learns the decomposition, but also implicitly learns the theoretically predicted behaviour on isolated eigenfunctions even if no isolated eigenfunctions were present in the training set. Hence the network \tg{generalises well and} is able to unlock the inherent structure of the non-linear spectral decomposition;
    \item We perform a comprehensive comparative study that shows the optimality of our architecture for non-linear spectral decomposition; 
    \item To the best of our knowledge, we are the first ones to propose a deep learning approach to approximate the non-linear spectral decomposition of images. 
\end{itemize}{}

\section{Background} \label{sec.background}
\subsection{Spectral Total Variation Decomposition}
Let $\Omega \subset \mathbb{R}^N$ be a bounded image domain with Lipschitz continuous boundary $\partial \Omega$. For an initial image $f:\Omega \to \mathbb{R}$ the total variation (TV) scale-space representation can be modelled by the gradient flow of $u:[0,\infty) \times\Omega \to \mathbb{R}$:
\begin{align}\label{eq.TVflow}
u_t(t;x) = -p(t;x), \,\, u(0;x) = f(x), \,\, p(t;x) \in \partial J_{TV}(u),
\end{align} 
where $\partial J_{TV}(u)$ denotes the subdifferential~\cite{rockafellar:1970} at $u$ of the following convex TV functional 
\begin{align}\label{eq.TV}
J_{TV}(u) = \sup_{\varphi} \left\{  \int_{\Omega} u \, \text{div}\, \varphi  \, dx, \lvert \varphi \rvert_{L^{\infty}} \leq 1 \right\} = \int_{\Omega} \lvert Du \rvert,
\end{align}
where the supremum is taken over $\varphi \in C_0^1(\Omega; \mathbb R^N)$ and $\lvert \varphi \rvert_{L^{\infty}} = \sup_{x \in \Omega} \sqrt{\sum_{i = 1}^{N}\varphi^2_i(x)}$. The element $p\in \partial J_{TV}(u)$ in~\eqref{eq.TVflow} is a subgradient of the TV functional~\eqref{eq.TV} at $u$.
We refer to \eqref{eq.TVflow} as the TV flow \cite{Andreu2001}.

\begin{figure*}
	\centering
	\subfigure[Input $f$ \label{subfig.input_f}]{\includegraphics[width = 0.175\textwidth]{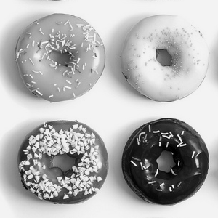}}
	\hspace{0.05cm}
	\subfigure[TV-Spectrum $S(t)$\label{subfig.spectrum}]{\includegraphics[width = 0.245\textwidth]{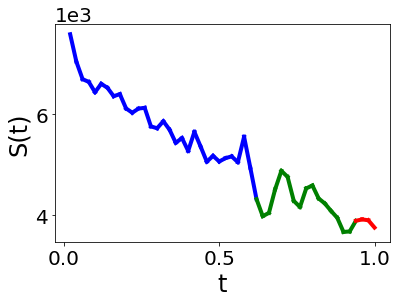}}
	\hspace{0.05cm}
	\subfigure[TV-High-pass filtered (blue)\label{subfig.phismall}]{\includegraphics[width = 0.17\textwidth]{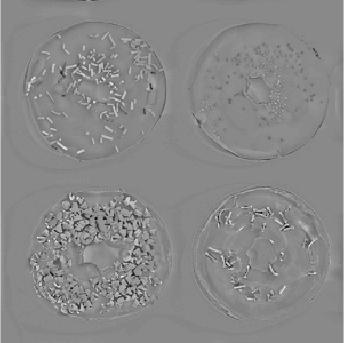}}
	\hspace{0.05cm}
	\subfigure[TV-Band-pass filtered (green)\label{subfig.phimed}]{\includegraphics[width = 0.17\textwidth]{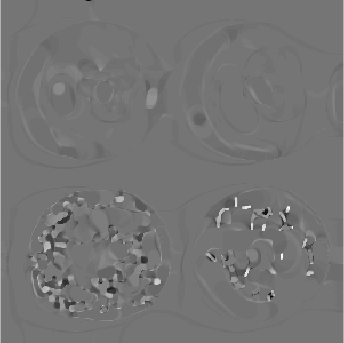}}
	\hspace{0.05cm}
	\subfigure[TV-Low-pass filtered (red)\label{subfig.philarge}]{\includegraphics[width = 0.17\textwidth]{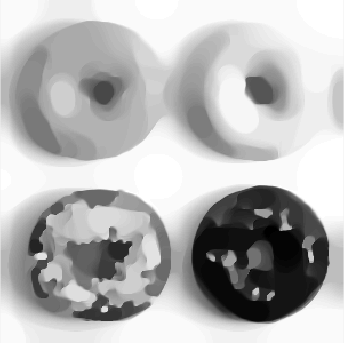}}
	\caption{Example of the filtered spectral responses of a natural image $f$ at different scales $t$. \subref{subfig.input_f} the initial image with \subref{subfig.spectrum} the spectrum $S(t)$. TV High- (blue), band- (green) and low-pass filtered (red) spectral bands depicting small to large structures \subref{subfig.phismall}--\subref{subfig.philarge} separating the sprinkles from the donuts. The input image is taken from MS COCO \cite{Lin2014}.
	\label{fig.spectrum}}
\end{figure*}
In order to decompose an image $f$ into its non-linear spectral components (in terms of the TV functional), \cite{Gilboa2014} introduced the TV transform that is defined using the solution of the TV flow $u$ by
\begin{align} \label{eq.TVtrafo}
	\phi(t;x) = u_{tt}(t;x)t,
\end{align}
where $u_{tt}$ denotes the second temporal derivative of $u$. 
The derivation should be understood in a weak sense (for the formal setting in spatial discrete and continuous domains see \cite{Burger2016} and \cite{bungert2019nonlinear}, respectively). 
The time parameter $t$ is also referred to as the \emph{scale} and plays a role analogous to the frequency in Fourier analysis. The TV transform $\phi(t;x)$ is invariant with respect to rotations and translations of the initial image $f$. However, spatial scaling and change of contrast lead to changes in the transform domain, that is structures will be recovered in different bands \cite{Gilboa2014}. 

For a spectral representation of an image, we generally expect the transform to generate impulses at some basic structures such as sines and cosines in the Fourier transform. In the case of total variation these basic structures are  functions $u$ satisfying $\lambda u \in \partial J_{TV}(u)$ with $\lambda \in \mathbb{R}$, which are also referred to as non-linear eigenfunctions ~\cite{Gilboa2018}. They create an impulse at scale $t = 1/\lambda$. Examples of such  eigenfunctions for the TV functional are multiples of the indicator function of a disk.  The scale at which an impulse is generated for the disk depends on the radius and the height of the disk. The spectrum of the TV transform, for $\phi \in L^1$,  is defined by 
\begin{align*}
S(t) = \lVert \phi (t;x) \rVert_{L^1(\Omega)} = \int_{\Omega} \lvert \phi (t;x) \rvert dx,
\end{align*}
and represents the $L^1$ amplitude of the spectral responses $\phi(t;x)$ at different scales $t$. An alternative definition which admits a Parseval-type equality is proposed in \cite{Burger2016}, for simplicity we will use the above definition. An example of the spectrum as well as the TV transform and the corresponding structures at different scales are shown in Figure \ref{fig.spectrum}. 

Given the spectral responses $\phi (t;x)$, the initial image $f$ of the TV flow can be recovered through the inverse transform defined by
\begin{align} \label{eq.inverse_trafo}
	f(x) = \int_{0}^{\infty} \phi(t;x) \,dt + \bar{f},
\end{align}
where $\bar{f}$ is the mean value of $f$. If we truncate the integral at some $T \in(0,\infty)$, we get
\begin{align} \label{eq.finite_inverse_trafo}
f(x) = \int_{0}^{T} \phi(t;x) \,dt + f_r(T;x),
\end{align}
where $f_r(T;x) = u(T;x)-u_t(T;x)T$ is referred to as the residual \cite{Gilboa2014}. 

For small scales $t$ the TV transform $\phi(t;x)$ consists of structures with small spatial size and low contrast in the initial image $f$. Coarser spatial features and those with higher contrast are contained in the spectral components $\phi(t;x)$ with larger scales $t$ (cf. Figure \ref{fig.spectrum}). 

This connection between scale and features can be used to manipulate features based on scale using a filter function $H: [0,\infty] \to \mathbb{R}$
\begin{align} \label{eq.filteredTVtrafo}
	\phi_H(t;x) = \phi(t;x)H(t).
\end{align}
Substituting the filtered TV transform \eqref{eq.filteredTVtrafo} into the inverse TV transform \eqref{eq.inverse_trafo} or \eqref{eq.finite_inverse_trafo} recovers the filtered image in the spatial domain. TV-band-pass filtering for scales in $[t_{k-1},t_k]$, $k = 1, \dots, K$, with $t_{k-1}<t_k $, in a finite time setting can be done using
\begin{align} \label{eq.spectral_bands}
    b^k = \int_{t_{k-1}}^{t_k} \phi(t;x) dt, \, \quad k = 1, \dots, K-1; \quad
    b^K = \int_{t_{K-1}}^{t_K} \phi(t;x) dt + f_r(t_K,x).
\end{align}
The resulting filtered images $\{b^1, \dots, b^K \}$ are referred to as bands and the initial image $f$ is then said to be decomposed into $K$ spectral bands, where for $t_0=0$, using \eqref{eq.finite_inverse_trafo}, the following identity holds: $f = \sum_{k=1}^K b^k$.

\subsection{Numerical Solution of the TV flow }\label{sec.classical}
One approach to numerically compute a spectral TV decomposition of an image $f$ is to discretise the TV flow \eqref{eq.TVflow}.  This is challenging due to the strong non-linearity and singularities in the subgradient $p$. A classical approach in the literature uses a regularised version of $p$ which reads $\mathrm{div}\left(Du/\sqrt{\lvert Du \rvert^2+\varepsilon}\right)$, and uses finite differences and appropriate time-stepping for numerical approximation \cite{chan1999convergence}. Here, explicit Euler time-stepping is computationally unstable, demanding a prohibitively small timestep size. More recent approaches approximate the TV flow by an implicit Euler scheme, which, as it turns out, can be realised via the following optimisation problem \cite{Gilboa2014, Burger2016}
\begin{align} \label{eq.ROF}
u(t+dt) = \argmin_{v} \frac{1}{2} \lVert u(t)-v \rVert_{L^2}^2 + dt\,J_{TV}(v),
\end{align}
where $u(t)$ is the solution of the gradient flow at time $t$ and $dt>0$  is the (sufficiently small) step size. The price to pay for numerical stability of the implicit Euler scheme is the non-smoothness of the optimisation problem  \eqref{eq.ROF}, which makes it computationally expensive, even with state-of-the-art SOTA convex optimisation techniques such as PDHG or ADMM \cite{chambolle2016introduction}. 

Once a solution of the TV flow~\eqref{eq.TVflow} has been obtained, the TV transform \eqref{eq.TVtrafo} can also be discretised by finite differences and the $K$ spectral TV bands of the initial image can then be recovered by filtering.  We note that such an accurate solution of the TV flow (and hence of spectral TV) requires solving \eqref{eq.ROF} $N$ times, for computing $N$ time steps of size $dt$. This strongly motivates the use of alternative faster approximations. We use decompositions obtained in this manner as the ground truth of our training.

\section{Deep Learning Approach to Spectral Decomposition}
While the decomposition of an image into its TV-spectral bands gives qualitatively highly desirable results, its computational realisation is cumbersome as it amounts to the solution of a series of non-smooth optimisation problems \eqref{eq.ROF}. For this reason, we propose a neural network approach for obtaining a spectral image decomposition. We show that our proposed TVspecNET can approximate \tg{the spectral} TV decomposition and can be computed several orders of magnitude faster than the model driven approach in Section~\ref{sec.classical}, cf. also Figure \ref{fig.visual_comp_leo}.

There are several ways in which NNs can achieve a speed up in the spectral decomposition pipeline. Solving the TV flow is computationally the most expensive part. A great deal of work has recently been done on training NNs to solve PDEs, e.g. \cite{Berg2018, Sirignano2018, Bar2019, Lu2019, Raissi2019, Jagtap2020}. While they are able to achieve good results, they require the calculation of the PDE explicitly in the loss functional. This, however, would involve an explicit expression of the subgradient $p$ of TV in \eqref{eq.TVflow}, which is not desirable (cf. Section~\ref{sec.classical}). Therefore, we choose to replace the whole pipeline by a NN and learn the decomposed images directly from the initial image. In that, the network has to learn the PDE only implicitly. Our problem is formulated as follows.

Given a training set of images $\{f_i\}_{i=1}^N$ and their spectral bands $\{b_i^1, \dots, b_i^K\}_{i=1}^N$ as defined in \eqref{eq.spectral_bands}, we seek to find a neural network $\Psi(\cdot,\Theta)$ with learnable parameters $\Theta$ such that 
\begin{align*}
    \Psi(f_i,\Theta) \approx \{b_i^1, \dots, b_i^K\}.
\end{align*}

A particular challenge of this problem setting is that we expect our NN to give a higher dimensional output, that is multiple spectral bands, from a single image input.

Convolutional neural networks (CNNs) have proven to be a powerful tool in many image analysis applications, e.g. \cite{Ronneberger2015,Zhang2017,He2017}. This is due to their ability to leverage local neighbourhood information of pixels through convolutions. Some encoder-decoder type architectures such as the U-Net \cite{Ronneberger2015} that was originally designed for image segmentation, have been successfully applied in other decomposition tasks such as intrinsic image decomposition \cite{Janner2017,Li2018,Lettry2018,Yuan2019}. However, since the TV transform depends on the interplay between spatial scale and contrast, encoder-decoder type architectures may not be well suited for spectral TV. These and other architectures that use down-/upsampling transform the spatial scale without changing contrast and hence hinder their interplay. Therefore, we use the denoising convolutional neural network (DnCNN) \cite{Zhang2017}, which does not rely on down- and upsampling, as the basis of our network architecture. \tg{The DnCNN is a non-contracting feed-forward neural network that leverages network depth (i.e. multiple sequential layers) and batch normalisation together with residual learning to separate the noisy part of an image from the clean image.} Our TVspecNET consists of $L = 17$ sequential convolutional layers $h_l$, $l = 1, \dots, L$ with a rectified linear unit (ReLU) \cite{Krizhevsky2012} activation, i.e. $\sigma(a) = \max(0,a)$. The layer can be defined by
\begin{align*}
    h_{l+1} = \sigma(w_{l+1}*h_l + b_{l+1}), \quad l = 1,\dots, L-1,
\end{align*}
where $\Theta_L = \{w_{l},b_{l}\}_{l=1}^L$ are the trained convolution kernels of size $3 \times 3$ and the biases. For $l=1,...,L-1$, each layer consists of 64 channels. The number of output channels in the last layer corresponds to the number of decomposed bands. This deep network design has a receptive field of $35\times35$ pixels and is therefore able to recover larger features. 

Let $\{b^1(\Theta), \dots, b^K(\Theta)\}$ be the output of the network $\Psi(f,\Theta)$ for input image $f$, and $\{\hat{b}^1, \dots, \hat{b}^K\}$ the ground truth bands. To train our network, we use the normalised mean squared error (MSE) loss:
\begin{align} \label{eq.MSEloss}
    \mathcal{L}(\Theta) = \frac{1}{K}\sum_{j = 1}^{K}\frac{\lVert b^j(\Theta) - \hat{b}^j\rVert_2^2}{\lVert \hat{b}^j \rVert_2^2}.
\end{align}
It is essential to use normalisation across the different bands in~\eqref{eq.MSEloss} to ensure that all bands contribute equally and bands with larger intensity ranges do not dominate the loss functional.

\section{Experiments}
In this section, we describe the experiments we conducted to evaluate the performance of our proposed TVspecNET. We describe the experimental setup and the evaluation scheme and demonstrate the performance of the network. We also perform a comprehensive architecture comparison to analyse the performance that a down-/upsampling based architecture could achieve. Additionally, we provide an ablation study in the Appendix to investigate different loss functionals.

\begin{table}
  \caption{Evaluation of the proposed TVspecNET on a testing dataset \cite{Lin2014} of 1000 images against the  model driven approach \cite{Gilboa2014} (cf. Section \ref{sec.classical}). Values correspond to averages over the dataset.}
  \label{tb.evaluation_measures}
  \centering
  \begin{tabular}{cccccccc}
    \toprule
          & Average     & Band 1 & Band 2 & Band 3 & Band 4 & Band 5 & residual Band \\
    \midrule
    SSIM & \textbf{0.9600}  & 0.9972  & 0.9906 & 0.9736 & 0.9441 & 0.8775 & 0.9771  \\
    PSNR     & \textbf{30.867} & 28.37 & 28.83 & 29.19 & 29.73 & 29.91 & 39.17    \\
    sLMSE     & \textbf{0.829} & 0.797 & 0.812 & 0.811 & 0.799 & 0.759 & 0.998 \\
    \bottomrule
  \end{tabular}
\end{table}

\subsection{Dataset and Training Settings}
For training and testing our neural network we use the MS COCO dataset \cite{Lin2014} that contains a large number of natural images. We take 2000 images for training and 1000 for testing. Each image is turned to greyscale and randomly cropped to a $64 \times 64$ pixel window. For the purpose of data augmentation, we also take $128 \times 128$ crops for some images and downsample them by a factor of 2, obtaining again images of size $64 \times 64$. As spectral TV decomposition is not invariant to cropping and resizing, this augmentation needs to be done during the data generation process and cannot be automated during training. After standardising the dataset to have zero mean and a standard deviation of 1, we generate $K = 50$ ground truth bands \eqref{eq.spectral_bands} using the  model driven approach in Section~\ref{sec.classical}. The bands are then combined dyadically to form 6 spectral bands. In this way, we make sure that smaller structures are decomposed in great detail while larger structures are grouped together in higher bands.

We train our network only for the first 5 bands, since the 6\textsuperscript{th} band contains the residual $f_r$ as described in \eqref{eq.spectral_bands} and can be recovered by subtracting the sum over bands 1-5 from the initial image. We use  the Adam optimiser \cite{Kingma2015} with an initial learning rate of $10^{-3}$ and multi step learning rate decay. Our neural network is trained with a batch size of 8 and for 5000 epochs on an NVIDIA Quadro P6000 GPU with 24 GB RAM.

\subsection{Evaluation Protocol} \label{sec.eval_protocol}
\begin{table}
  \caption{Properties of the TV transform evaluated for the TVspecNET on a testing dataset of 1000 images. The comparison is made between the TVspecNET output and the expected decomposition based on one-homogeneity, and translational and rotational invariance, respectively. The values correspond to averages over the dataset.}
  \label{tb.TV_properties}
  \centering
  \begin{tabular}{cccc}
    \toprule
          & one-homogeneity & translation invariance & rotation invariance \\
    \midrule
    SSIM & 0.9867  & 0.9930 & 0.9807 \\
    PSNR     & 33.783 & 37.593 & 32.042  \\
    sLMSE     & 0.880 & 0.983 & 0.885  \\
    \bottomrule
  \end{tabular}
\end{table}

We evaluate the performance of the network in three ways. Firstly, we give a quantitative evaluation on a testing set of 1000 natural images from the MS COCO dataset in terms of three common image quality measures, the structural similarity index (SSIM), the peak signal-to-noise-ratio (PSNR) and the inverted localised mean squared error (sLMSE) \cite{Grosse2009}. While the first two metrics are commonly used in image analysis, the sLMSE is more often found in the evaluation of intrinsic image decomposition and derives the local MSE on patches. Both for sLMSE and SSIM, the similarity between two images is high if the value is close to 1.

Secondly, we investigate whether the network is able to learn the underlying properties of the non-linear spectral decomposition that are predicted by the theory, such as one-homogeneity, and translational and rotational invariances. One-homogeneity implies that changing the contrast by a constant factor should shift the spectrum. For instance, if we multiply images by 2, we expect all image structures to appear in subsequent bands (since we use dyadic bands). Translational and rotational invariances imply that the spectral bands of a translated/rotated image experience the same translation/rotation as the original image. 

To further investigate how well the network can 'understand' the TV transform, we test whether a network trained on natural images can learn eigenfunctions of the TV transform, i.e. whether it demonstrates the predicted behaviour on isolated eigenfunctions (which are very different from the images in the training set). For this purpose, we test the performance of the network on disk images. 

\subsection{Results}
Firstly, we evaluate the performance of TVspecNET against ground truth decompositions obtain by solving a gradient flow (cf. Section \ref{sec.classical}) following the protocol in Section \ref{sec.eval_protocol}. A visual comparison for an example image is shown in Figure \ref{fig.visual_comp_leo}. Our network is able to recover all spectral TV bands almost perfectly. This is confirmed by quantitative measures of similarity (SSIM, PSNR and sLMSE) between TVspecNET decompositions and the ground truth, as shown in Table \ref{tb.evaluation_measures}. 

Secondly, we demonstrate that the trained network retains properties of the TV transform predicted by the theory: one-homogeneity and translational and rotational invariances. To test translational and rotational invariances, we apply rotations/translations to the original image and then apply the network or we apply the network and then translate/rotate the bands. If the results are the same, the network has the desired invariances. To test one-homogeneity, we multiply the initial image by a factor of 2 and compare each band of the scaled image with the previous band of the original image. Since we use dyadic bands, these bands should be the same (up to the multiplication factor of 2). Quantitative results in Table \ref{tb.TV_properties} demonstrate an almost perfect match in all three tests. While translational invariance is inherent to fully convolutional NNs and rotational invariance is to some extent enforced through data augmentation, one-homogeneity is neither explicitly enforced nor is the network penalised to retain this property.

\begin{figure}
    \centering
    \includegraphics[width=\textwidth]{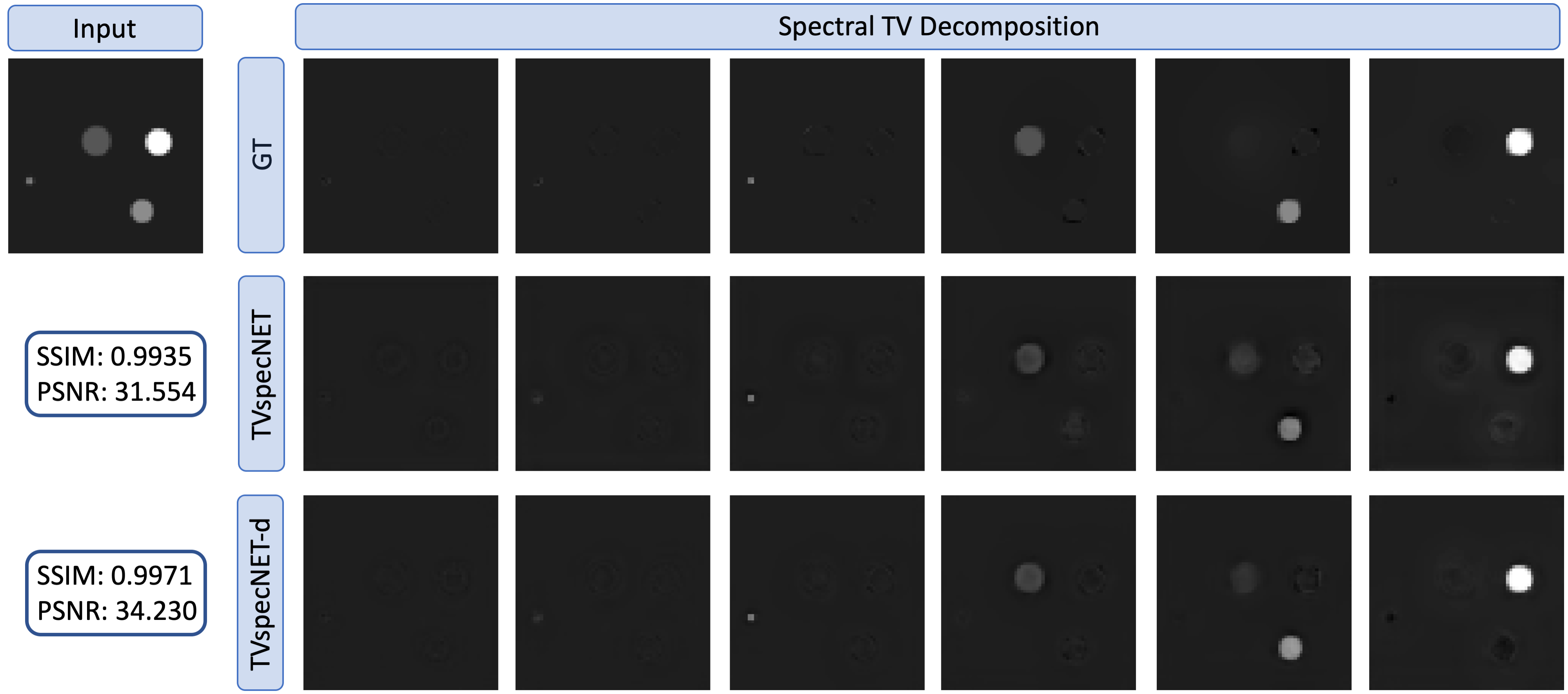}
    \caption{Visual comparison of a ground truth (GT) decomposition for a disk image and the output of TVspecNET trained only on natural images and TVspecNET-d that was additionally trained on disk images. Adding disks to the training set improves performance only slightly; even without seeing disks in the training set the network is able to learn the correct behaviour.
    }
    \label{fig.diks}
\end{figure}

Thirdly, we evaluate the performance of the network on images of isolated eigenfunctions. For this test, we create images of disks with various radii and contrasts. To see how well the network learns eigenfunctions from a dataset of natural images, we make a comparison with the same architecture trained on a dataset containing both natural images and images of isolated eigenfunctions (disks). For ease of distinction, we call the second network TVspecNET-d.
The results are shown in Figure \ref{fig.diks}. Remarkably, TVspecNET is capable of recovering spectral bands of disks (i.e. separating different disks into different bands) with high accuracy even without having seen them in the training set; adding eigenfunctions to the training set improves the performance only slightly. Similar observations can be made on  images of ellipses. We provide further visual and quantitative examples in the Appendix. 
Interestingly, the performance of TVspecNET-d on natural images is slightly worse compared to TVspecNET (TVspecNET-d has SSIM 0.9533, PSNR 30.622 and sLMSE 0.784). 

\begin{table}
  \caption{Computation time (in seconds) of the  model driven approach evaluated on a CPU (Matlab) and on a GPU (C++/Python), and of TVspecNET with three different basis networks evaluated on a GPU. Values correspond to averages over the dataset.}
  \label{tb.computation_time}
  \centering
  \begin{tabular}{cccccc}
    \toprule
    &\multicolumn{5}{c}{Total number of pixels per image}                   \\
    \cmidrule(r){2-6}
          & 4096 & 16384 & 65536 & 262144 & 1048576 \\
    \midrule
    Model Driven on CPU & 1.6341  & 3.8431  & 12.4277 & 74.3933 & 344.6169   \\
    Model Driven on GPU & 5.2104 & 5.2787 & 5.9412 & 7.8306 & 28.2811 \\
    TVspecNET    & \textbf{0.0020} & \textbf{0.0021} & \textbf{0.0021} & \textbf{0.0020} & \textbf{0.0020}\\
    F-TVspecNET     & \textbf{0.0020} & \textbf{0.0021} & 0.0022 & 0.0023 & 0.0024\\
    U-TVspecNET     & 0.0057 & 0.0055 & 0.0057 & 0.0059 & 0.0062\\
    \bottomrule
  \end{tabular}
\end{table}

Finally, we compare the computation time of TVspecNET and the  model driven approach in Section \ref{sec.classical} \tg{and show the results in Table \ref{tb.computation_time}}. For the  model driven approach we use two implementations, a Matlab implementation \cite{Gilboa2014} of the projection algorithm by Chambolle \cite{Chambolle2004} running on a CPU and a primal-dual implementation \cite{Chambolle2010,Hammernik2017} in C++/Python\footnote{Code used from \url{https://github.com/VLOGroup/primal-dual-toolbox}} running on a GPU. The NN is evaluated on the same NVIDIA Quadro P6000 GPU with 24 GB RAM. 
As expected, the GPU implementation of the  model driven approach is slower than the CPU implementation on small images (due to the GPU overhead), but becomes significantly faster for large images.
TVspecNET, however, is orders of magnitude faster than both implementations on all image sizes. For the largest image size we tested, $1024 \times 1024$ pixels, the speed up of TVspecNET compared to the GPU implementation of the  model driven approach is four orders of magnitude. 
Also, the computation time for the  model driven approach increases significantly with the number of pixels (due to the increased size of the optimisation problems) while for TVspecNET it remains approximately constant. 

\subsection{Comparison of Basis Architectures} \label{sec.ablation_study}
We compare different architectures as the basis of out NN: DnCNN \cite{Zhang2017} as proposed in TVspecNET,  FFDnet \cite{Zhang2018} (we call this network F-TVspecNET) and U-Net \cite{Ronneberger2015} (we call this network U-TVspecNET). While DnCNN does not rescale or downsample the image, both FFDnet and U-Net contain pooling to various extends. FFDnet downsamples the input to 4 low resolution images before applying the network and combines the results to form a high resolution denoised image. U-Net uses multiple maxpooling and upsampling steps within the network architecture.

\begin{table}
    \centering
    \caption{Comparison of different network architectures as the basis for our TVspecNET: DnCNN (TVspecNET), FFDnet (F-TVspecNET) and U-Net (U-TVspecNET).
    \label{tb.network_comparison}}
    \begin{tabular}{cccc}
         \toprule
             & TVspecNET & F-TVspecNET & U-TVspecNET  \\
             \midrule
             SSIM & \textbf{0.9600}  & 0.9377  & 0.9233\\
             PSNR     & \textbf{30.867} & 28.098  & 28.993\\
             sLMSE     & \textbf{0.829} & 0.6854  & 0.7382\\
             \bottomrule
    \end{tabular}
\end{table}

The results of this comparison are shown in Table \ref{tb.network_comparison}. The proposed TVspecNET clearly outperforms both F-TVspecNET and U-TVspecNET, confirming our choice of the basis architecture. Although downsampling used in FFDnet and U-Net increases the receptive field, which is useful to recover larger features, for non-linear spectral decompositions (TV)  downsampling turns into a disadvantage, since it hinders the interplay between size and contrast, which is crucial. In terms of computational time (Table \ref{tb.computation_time}), TVspecNET and F-TVspecNET are comparable while U-TVspecNET is approximately three times slower (still, all three are orders of magnitude faster than the  model driven method).

\section{Conclusion}
In this paper, we propose TVspecNET, a neural network that can learn a non-linear spectral decomposition. We show, both through  qualitative and quantitative analysis, that TVspecNET is able to decompose images into spectral TV bands. 
The most striking result is that, even though the network is trained only on natural images, it is able to learn basic structures of the non-linear TV transform (eigenfunctions) and its properties such as one-homogeneity and rotational/translational invariance. Without incorporating any explicit knowledge about the underlying PDE of the spectral decomposition (TV flow) into training, this data-driven learning approach is able to  learn the PDE implicitly. The speed-up that the network achieves is also impressive, going up to four orders of magnitude on $1024 \times 1024$ \tg{px} images compared the state-of-the-art GPU implementation of the model driven approach. \tg{The code for TVspecNET is publicly available on Github\footnote{\url{https://github.com/TamaraGrossmann/TVspecNET}}.}
An interesting direction for future work is 'inverting' the process and learning the decomposition from user-defined eigenfunctions, and applying the trained network to real images.

\begin{ack}
We thank Angelica I. Aviles-Rivero, Christian Etmann, Damian Kaliroff, Lihao Liu, Tomer Michaeli and Tamar Rott Shaham  for helpful discussions and advice.

This work was supported by the European Union’s Horizon 2020 research and innovation programme under the Marie Sk{\l}odowska-Curie grant agreement No. 777826 (NoMADS). TG, YK and CBS acknowledge the support of the Cantab Capital Institute for the Mathematics of Information. TG additionally acknowledges support from the EPSRC National Productivity and Investment Fund grant Nr. EP/S515334/1 reference 2089694. YK acknowledges the support of the Royal Society (Newton International Fellowship NF170045 Quantifying Uncertainty in Model-Based Data Inference Using Partial Order) and the National Physical Laboratory. GG acknowledges support by the Israel Science Foundation (Grant No.  534/19) and by the Ollendorff Minerva Center. CBS  acknowledges  support  from  the  Leverhulme  Trust  project  on  Breaking  the  non-convexity  barrier,  the  Philip  Leverhulme  Prize,  the  EPSRC  grants  EP/S026045/1  and EP/T003553/1,  the  EPSRC  Centre  Nr. EP/N014588/1,  European  Union  Horizon  2020 research  and  innovation  programmes  under  the  Marie  Sk{\l}odowska-Curie grant agreement No. 777826 NoMADS and No. 691070 CHiPS, and the Alan Turing Institute.

We also acknowledge the support of NVIDIA Corporation with the donation of two Quadro P6000, a Tesla K40c and a Titan Xp GPU used for this research.

\end{ack}

\bibliographystyle{abbrv}
\bibliography{bibliography}

\newpage
\appendix
\section*{Appendix}
\section{Outline}
In this appendix, we expand the results presented in the main paper primarily focussing on additional comparisons and new types of input images. In that, we show the optimality of our network in terms of loss functionals and highlight the generalisibility further. The appendix is structured as follows: First, we show visual result on disk, ellipse and natural images in Section \ref{sec.additional_results} as well as for one-homogeneity, translational and rotational invariance. In Section \ref{sec.ablation_study_supp} we perform an ablation study to compare the impact of different loss functionals on the results. Lastly, we give a formal definition of the quantitative metrics used for evaluation of our proposed TVspecNET in Section \ref{sec.eval_metrics}.

\begin{figure}
    \centering
    \subfigure[Decomposition of an image of a leopard. Different bands are scaled separately to better show the structures in the first bands. Image taken from the BSDS500 dataset \cite{Martin2001}.]{\includegraphics[width=\textwidth]{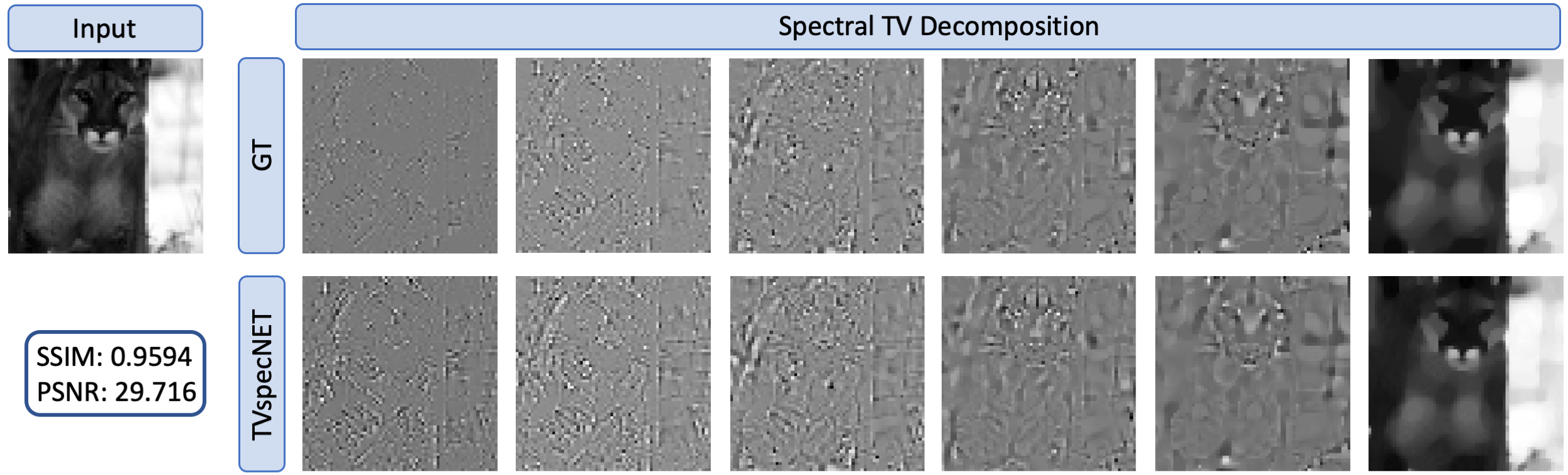}}
    \subfigure[Decomposition of an image of four donuts. All bands have the same scaling. Image taken from the MS COCO dataset \cite{Lin2014}.]{\includegraphics[width=\textwidth]{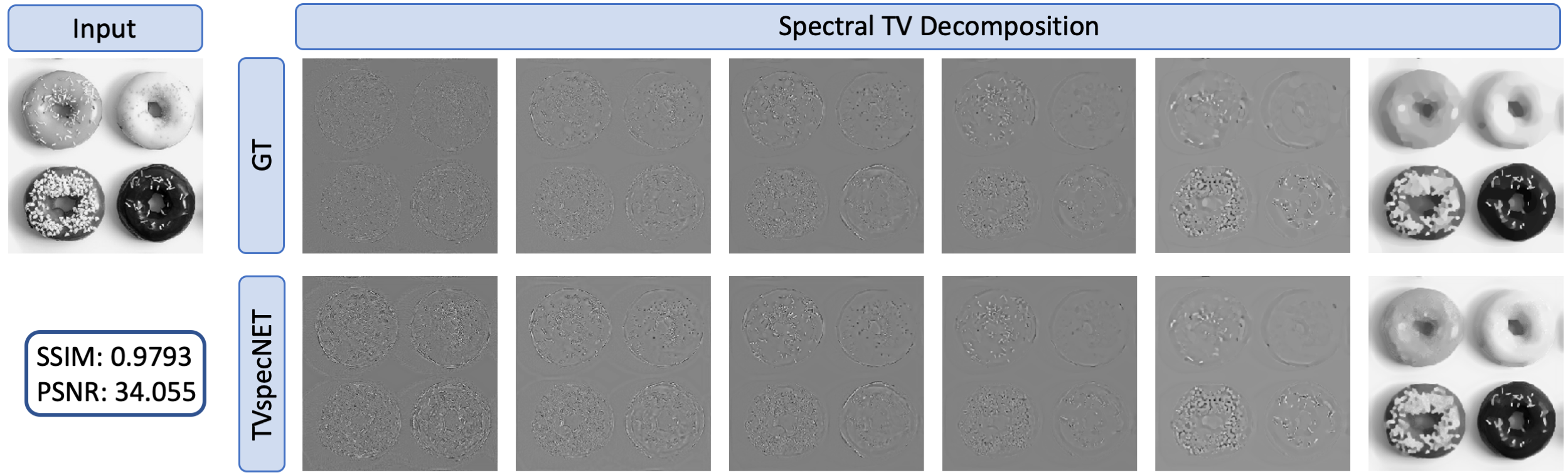}}
    \subfigure[Decomposition of an image of a giraffe. All bands have the same scaling. We show the difference between GT and TVspecNET for each band in the third row. Image taken from the MS COCO dataset \cite{Lin2014}.]{\includegraphics[width=\textwidth]{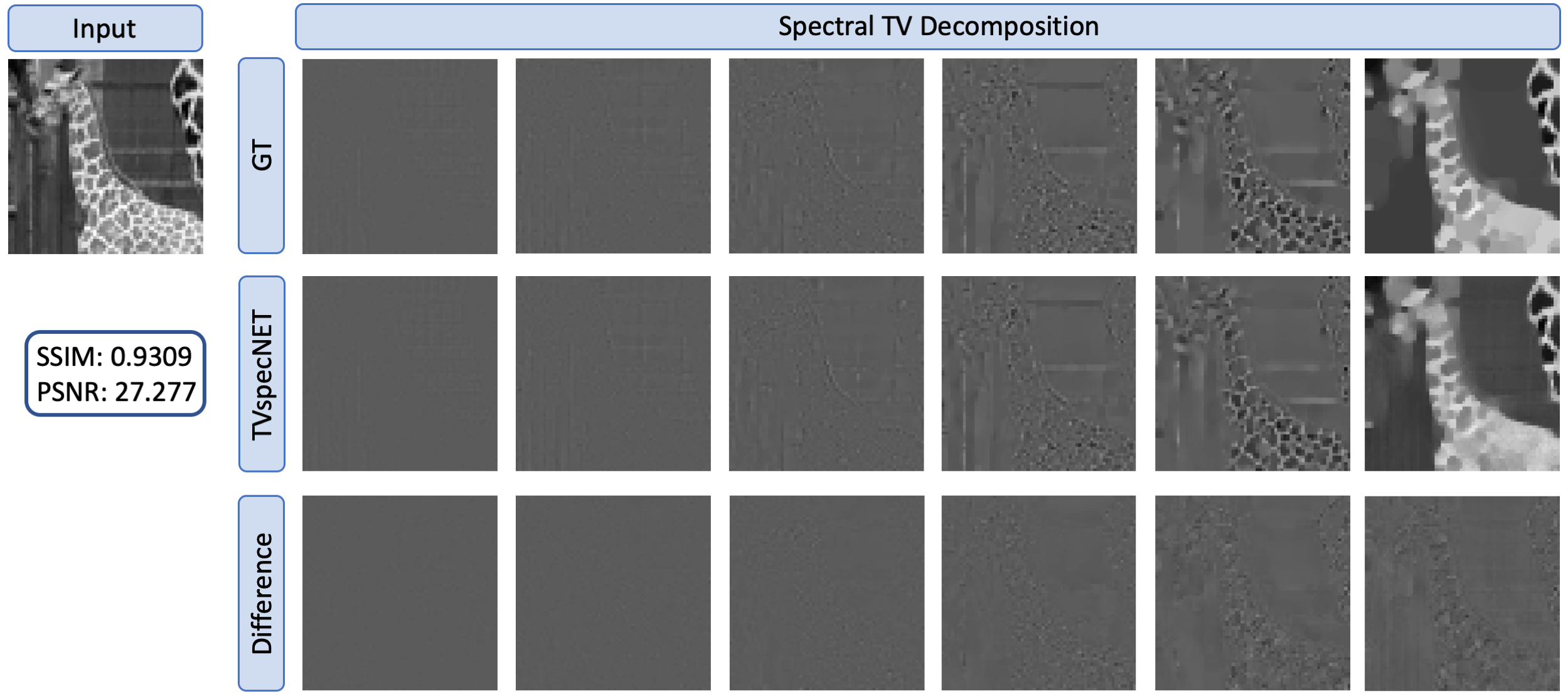}}
    \caption{Visual comparison of the TVspecNET decomposition and the ground truth (GT) \cite{Gilboa2014} on three natural images. In (a) we scale the bands individually to better show the structures in the first bands, whilst in (b) and (c) all bands have the same scaling. In (c) we additionally show the difference between GT and TVspecNET bands.}
    \label{fig.decomp_examples}
\end{figure}

\section{Additional Experimental Results} \label{sec.additional_results}
\begin{figure}
    \centering
    \includegraphics[width=\textwidth]{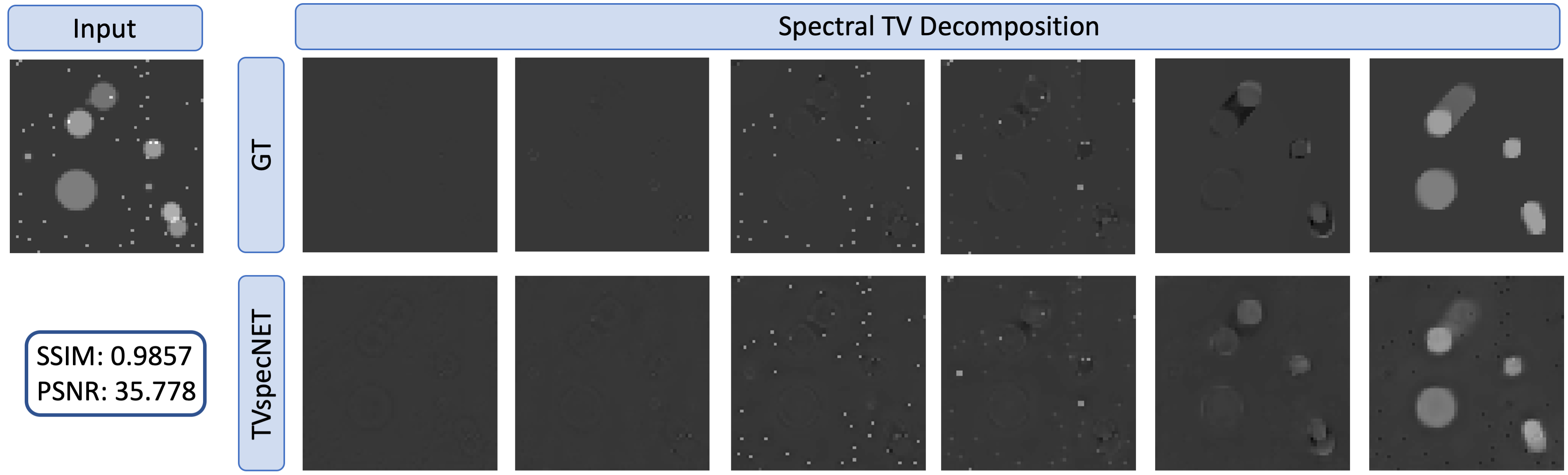}
    \caption{Visual comparison of the TVspecNET decomposition and the ground truth (GT) \cite{Gilboa2014} on an image containing multiple disks.}
    \label{fig.disks_decomp}
\end{figure}

\begin{figure}[h!]
    \centering
    \includegraphics[width=\textwidth]{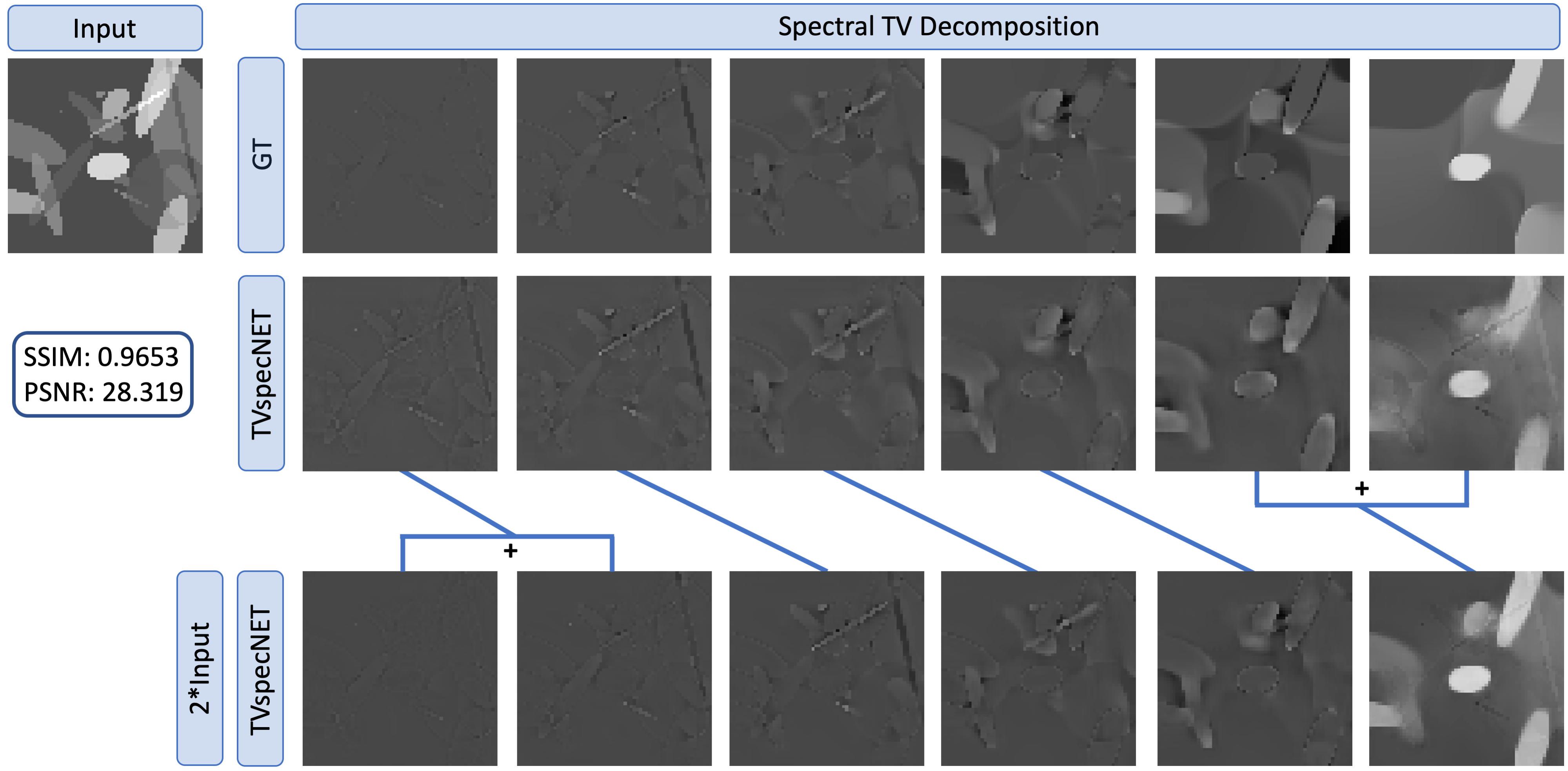}
    \caption{Visual comparison of the TVspecNET decomposition and the ground truth (GT) \cite{Gilboa2014} on an ellipse image (first two rows). The last row displays the TVspecNET results decomposition of the ellipse image multiplied by a factor of 2. The bands get shifted as expected from a TV spectral decomposition (corresponding bands are marked by lines). The contrast was normalised, i.e. the bands in the third row were divided by 2 to match the contrast of the first two rows. This example demonstrates the one-homogeneity property of TVspecNET. The first band is split into two bands when the input is multiplied by 2. Since the last band depicts the residual, the fifth and sixth bands from the original input are combined in the residual of the decomposition of the modified image.}
    \label{fig.ellipse}
\end{figure}

In addition to the results presented in the main paper, we give multiple visual examples of the TVspecNET decomposition on natural images compared to the ground truth in Figure \ref{fig.decomp_examples} and on a disk image in Figure \ref{fig.disks_decomp}. Moreover, we consider visual and quantitative results for images of ellipses. These images neither contain isolated eigenfunctions as in disk images that we considered earlier (although isolated ellipses are eigenfunctions of the TV transform), nor are they similar to the natural images; therefore these images are somewhere in between the two types of images considered in the main paper. We generated a small dataset of 20 images that contain multiple ellipses of different size and contrast using the Python toolbox ODL (Operator Discretization Library)\footnote{Documentation can be found at \url{https://odlgroup.github.io/odl/}}. On this dataset, TVspecNET achieves high quantitative performance measures: SSIM is 0.9658, PSNR is 30.609 and sLMSE is 0.728, which is similar to the performance on natural images (cf. Table \ref{tb.evaluation_measures} in the main paper). We show the visual results for an example ellipse image in Figure \ref{fig.ellipse}.

\begin{figure}[th]
    \centering
    \subfigure[Example on a randomly generated disk image.\label{fig.disk-1-hom}]{
    \includegraphics[width=\textwidth]{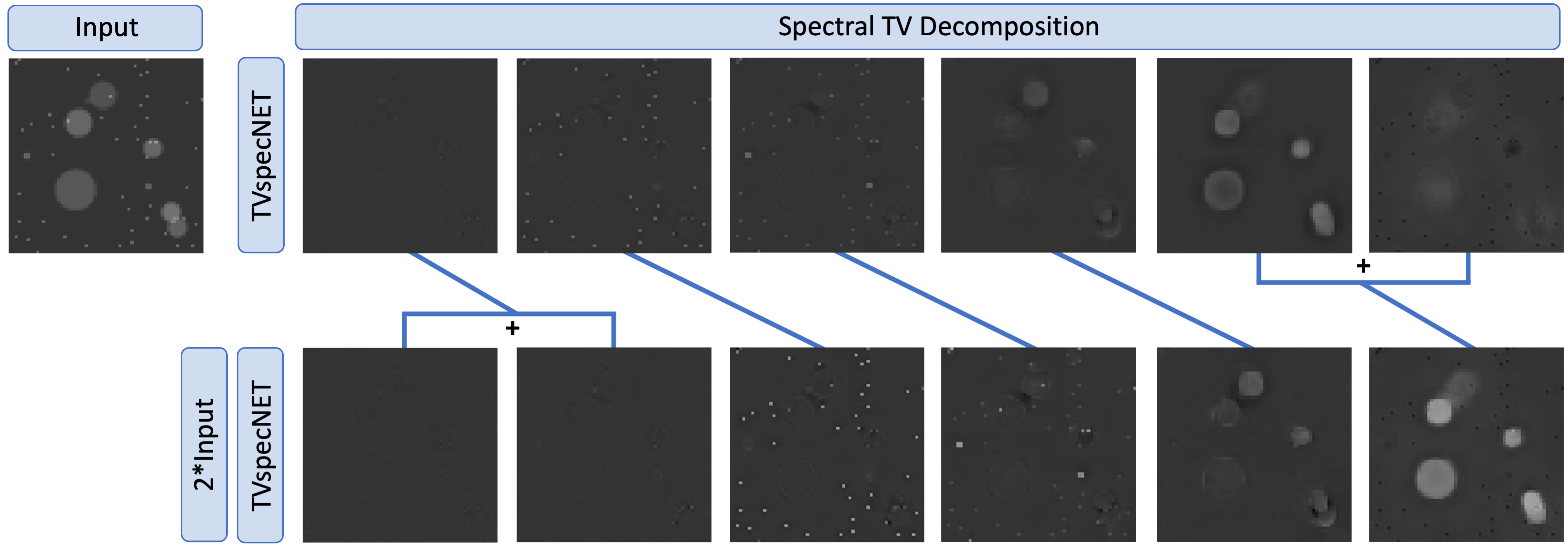}}
    \subfigure[Example on natural image from the MS COCO dataset \cite{Lin2014}. The contrast was normalised in this example, i.e. the bands in the second row were divided by 2 to match the contrast of the first row. \label{fig.natural-1-hom}]{\includegraphics[width=\textwidth]{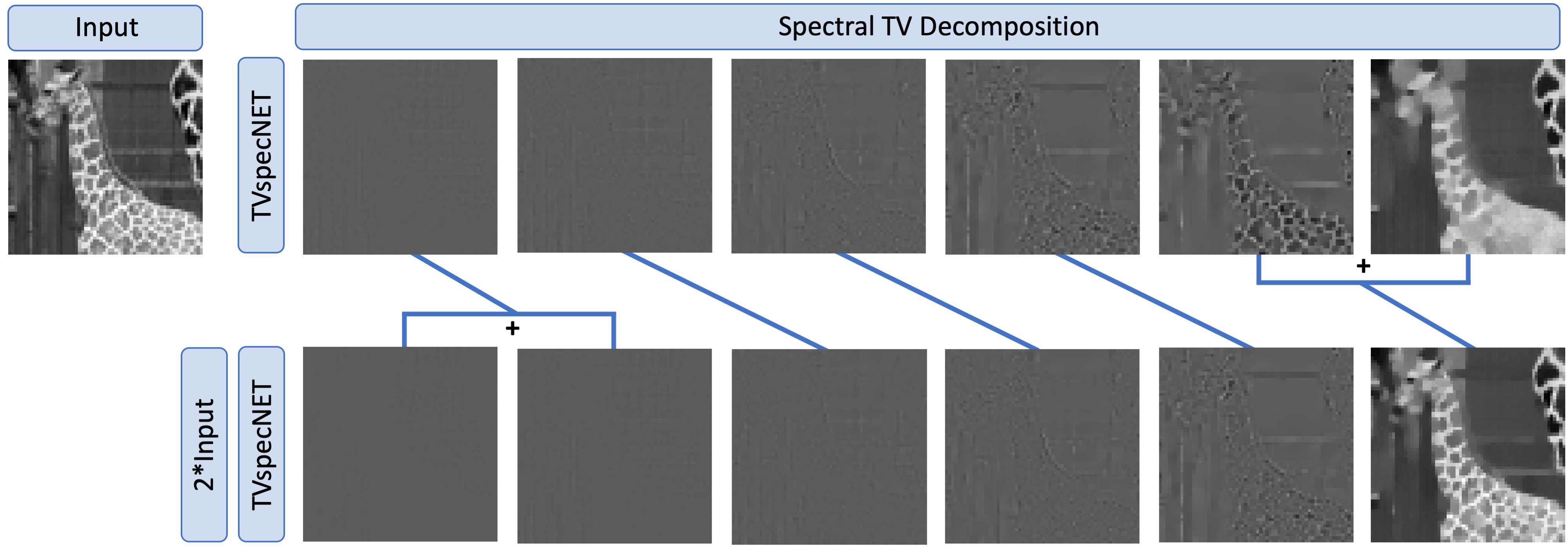}}
    \caption{Visual demonstration of the one-homogeneity property on (a) a disk image and (b) a natural image. In each case, the lower row displays TVspecNET decomposition of the image in the upper row multiplied by a factor of 2. Corresponding bands are marked by lines. The first band of the original image is split into two bands when the image is scaled  by a factor of 2. Since the last band contains the residual, the fifth and sixth bands from the decomposition of the original image will be combined in the residual of the scaled input.}
\end{figure}

Furthermore, we visually demonstrate one-homogeneity of TVspecNET for all image types we have so far presented. We multiply an input image by a factor of 2 and expect the resulting bands to be the same as in the original input image up to a shift into the next dyadic band (and an increase of contrast by a factor of 2). For an example image containing disks, we show the comparison between the spectral TV decomposition of the two input images obtained with TVspecNET in Figure \ref{fig.disk-1-hom}. The one-homogeneity property for an ellipse image is demonstrated in Figure \ref{fig.ellipse} in the two bottom rows. Finally, for a natural image we show the same property in Figure \ref{fig.natural-1-hom}. Overall,  we observe that one-homogeneity holds true for all image types considered in this work. 

\begin{figure}[th]
    \centering
    \subfigure[The input image (top) is translated (bottom) to show translational invariance. The spectral bands displayed are cropped to the area of interest, corresponding to a back-translation in the bottom row.]{
    \includegraphics[width=\textwidth]{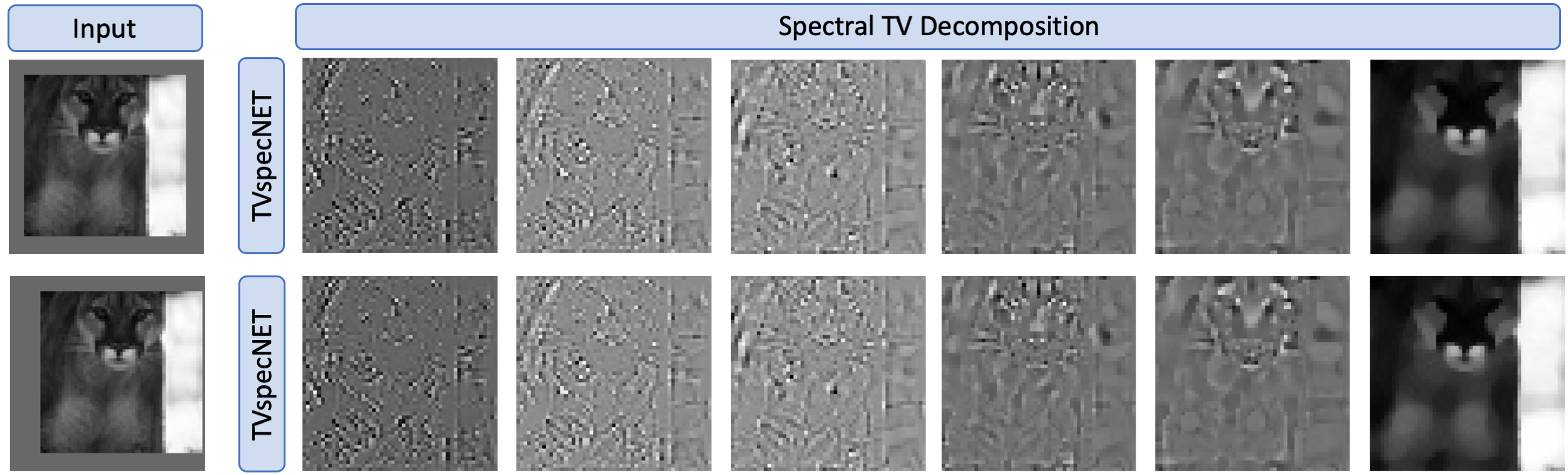}}
    \subfigure[The input image (top) is rotated (bottom) to show rotational invariance.]{\includegraphics[width=\textwidth]{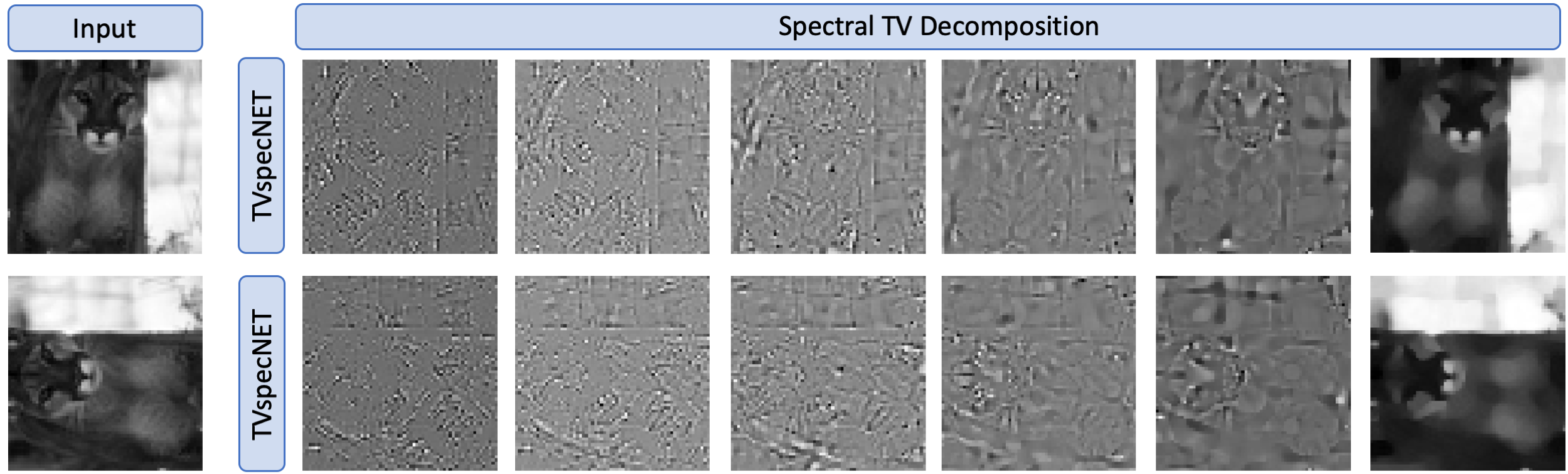}}
    \caption{Visual demonstration of translational and rotational invariances of TVspecNET on a natural image from the BSDS500 dataset \cite{Martin2001}.}
    \label{fig.trans_rot_comparison}
\end{figure}

Lastly, we give example results for translational and rotational invariances of TVspecNET on a natural image in Figure \ref{fig.trans_rot_comparison}. Spectral bands of a rotated/translated image are expected to be the same as in the original image up to the same rotation/translation. 

\tg{As an extension, we trained the same neural network for a finer graded decomposition with 25 bands and the results are similarly convincing (SSIM: 0.958, PSNR: 30.228) as for dyadic bands. However, it is not necessary to retrain the network for a larger number of bands in order to obtain a finer graded decomposition of certain scales: due to one-homogeneity one can shift bands in an image by multiplying it by a constant factor (factor <1 will shift larger structures to smaller bands, a factor >1 will do the reverse). Hence, we can represent the structures of interest at a finer scale through shifting without having to retrain the network.}

\section{Ablation Study}\label{sec.ablation_study_supp}
In the proposed TVspecNET, we employ the mean squared error (MSE) as the loss functional (denoted by $\mathcal{L}$). Considering losses that additionally model or penalise specific properties of the TV transform may be beneficial to recovering the spectral bands. Therefore, we investigate whether more complex loss functionals are able to recover the spectral bands at a higher image quality. Let $\{b^1(\Theta), \dots, b^K(\Theta)\}$ be the network output and $\{\hat{b}^1, \dots, \hat{b}^K\}$ the corresponding ground truth spectral bands for an input image $f$. As the spectral TV decomposition is edge-preserving, we include the normalised Huber loss of the image gradients,
\begin{align} \label{eq.Gradloss}
    \mathcal{L_{r}}(\Theta) = \frac{1}{K}\sum_{j = 1}^{K}\frac{\lVert \nabla b^j(\Theta) - \nabla \hat{b}^j\rVert_{Huber}}{\lVert \nabla \hat{b}^j \rVert_{Huber}},
\end{align}
to align edges in the bands. Furthermore, using the inverse TV transform introduced in the main paper, we enforce that the sum over all bands is equal to the input image:
\begin{align} \label{eq.Sumloss}
    \mathcal{L_{\sum}}(\Theta) = \frac{\lVert\sum_{j = 1}^{K} b^j(\Theta) - f \rVert_2^2}{\lVert f \rVert_2^2}.
\end{align}
We train four networks with the same parameter and training settings, changing only the loss functionals; we use the MSE loss $\mathcal{L}$ as well as its combinations with the losses introduced in~\cref{eq.Gradloss,eq.Sumloss}:
\begin{equation*}
\mathcal L_1 = \mathcal{L} + \mathcal{L}_{\sum}, \quad  \mathcal L_2 = \mathcal{L} + \mathcal{L_{r}}, \quad \mathcal L_3 = \mathcal{L} + \mathcal{L}_{\sum} + \mathcal{L_{r}}.    
\end{equation*}

The quantitative results for the ablation study are shown in Table \ref{tb.ablation_total}. We observe no significant improvement in performance compared to the MSE loss. While $\mathcal{L}_1$ has the highest SSIM and PSNR, any variations between losses are very slight. This confirms that the influence of a more complex loss functional on the decomposition retrieval is very limited. The simpler MSE loss is therefore able to recover the spectral TV features and properties without explicitly including this knowledge on TV transform properties at the same high image quality.

\begin{table}[th] 
    \centering
    \caption{Comparison of different loss functionals (cf. \cref{eq.Gradloss,eq.Sumloss}) with $\mathcal{L}$ the MSE loss: $\mathcal L_1 = \mathcal{L} + \mathcal{L}_{\sum}$, $\mathcal L_2 = \mathcal{L} + \mathcal{L_{r}}$, $\mathcal L_3 = \mathcal{L} + \mathcal{L}_{\sum} + \mathcal{L_{r}}$. Adding complexity to the loss does not improve performance significantly. \label{tb.ablation_total}}
        \centering
        \begin{tabular}{ccccc}
            \toprule
            & $\mathcal{L}$ & $\mathcal L_1$ & $\mathcal L_2$  & $\mathcal L_3$ \\
            \midrule 
                SSIM & 0.9600  & \textbf{0.9639}  & 0.9559 & 0.9619  \\
                PSNR     & 30.867 & \textbf{30.889} & 30.714 & 30.872  \\
                sLMSE     & \textbf{0.8290} & 0.8260 & 0.8224 & 0.8260 \\
            \bottomrule
        \end{tabular}
\end{table}

\section{Evaluation Metrics} \label{sec.eval_metrics}
For readers' convenience, we give explicit definitions of the evaluation metrics used throughout the main paper and the Appendix. Let $b$ be an output band of the TVspecNET and $\hat{b}$ the corresponding ground truth band. We evaluate the performance of the TVspecNET on each band individually and average over all bands to obtain an overall image quality value. The three evaluation metrics we use are defined as follows.

\paragraph{PSNR} The peak signal-to-noise ration (PSNR) is derived from the MSE between images on a logarithmic scale as follows:
\begin{align*}
    \text{PSNR}(b,\hat{b}) = 10*\log_{10}\left( \frac{\text{MAX}_I^2}{\text{MSE}(b,\hat{b})} \right),
\end{align*}
where $\text{MAX}_I$ is the maximal possible intensity value for spectral TV bands. Higher PSNR values correspond to a larger signal-to-noise ratio and therefore a better recovery of the ground truth bands.

\paragraph{SSIM} The structural similarity index measure (SSIM) \cite{Wang2004} describes the similarity between two images based on differences in luminance, contrast and structure. For $\mu_b, \mu_{\hat{b}}$ the mean intensities of images $b, \hat{b}$ and $\sigma_b, \sigma_{\hat{b}}$ their standard deviations, the SSIM is defined as:
\begin{align*}
    \text{SSIM}(b,\hat{b}) = \frac{(2\mu_b \mu_{\hat{b}} + c_1)(2\sigma_{b \hat{b}}+c_2)}{(\mu_b^2 +  \mu_{\hat{b}}^2 + c_1)(\sigma_b^2 +  \sigma_{\hat{b}}^2 + c_2)},
\end{align*}
where $\sigma_{b \hat{b}}$ denotes the covariance of $b$ and $\hat{b}$, and $c_1, c_2$ are constants that avoid a blow-up when the denominator is small. The SSIM value increases with larger similarity between images and a takes value of 1 for perfect approximation.

\paragraph{sLMSE} The inverted localised mean squared error (sLMSE)~\cite{Grosse2009} has been developed to soften the strict MSE metric. Based on the local MSE (LMSE) on image patches, the sLMSE prevents localised errors from dominating the overall image error. For patches $b_{\omega},\hat{b}_{\omega}$ of size $k \times k$ for some estimated and ground truth images $b$ and $\hat{b}$, the sLMSE is defined as
\begin{align*}
    \text{LMSE}(b,\hat{b}) = \sum_{\omega} \lVert b_{\omega} - \hat{b}_{\omega} \rVert_2^2, \quad sLMSE(b,\hat{b}) = 1 - \frac{LMSE(b,\hat{b})}{LMSE(0,\hat{b})}.
\end{align*}
In our case, we choose $k = 16$ with a step size of $8$ pixels. sLMSE is an inverted measure, meaning that two images have high similarity if the sLMSE value is close to 1.

\end{document}